\documentclass[lettersize,journal]{IEEEtran}
\usepackage{amsmath,amsfonts}
\usepackage{algorithmic}
\usepackage{algorithm}
\usepackage{array}
\usepackage[caption=false,font=normalsize,labelfont=sf,textfont=sf]{subfig}
\usepackage{textcomp}
\usepackage{stfloats}
\usepackage{url}
\usepackage{verbatim}
\usepackage{graphicx}
\usepackage{cite}
\usepackage{relsize}
\usepackage{graphicx}
\usepackage{newfloat}
\usepackage{listings}
\DeclareCaptionStyle{ruled}{labelfont=normalfont,labelsep=colon,strut=off} 
\floatstyle{ruled}
\newfloat{listing}{tb}{lst}{}
\floatname{listing}{Listing}
\usepackage{graphicx}
\usepackage{amsmath}

\usepackage{amssymb}
\usepackage{booktabs}

\usepackage{color}
\newcommand{\tablestyle}[2]{\setlength{\tabcolsep}{#1}\renewcommand{\arraystretch}{#2}\centering\footnotesize}
\usepackage[colorlinks,
            linkcolor=blue,
            anchorcolor=blue,
            citecolor=blue]{hyperref}

\begin{document}


\title{CusEnhancer: A Zero-Shot Scene and Controllability 
Enhancement Method for Photo Customization via  ResInversion} 

\author{Maoye Ren, Praneetha Vaddamanu, Jianjin Xu, Fernando De la Torre Frade
\thanks{Maoye Ren are with the College of Computer Science and
Technology, East China University of Science and Technology, Shanghai 200237, China. This work was finished at Carnegie Mellon University}
\thanks{Jianjin Xu and Fernando De la Torre Frade are with the College of Computer Science, Carnegie Mellon University,  Pittsburgh, 15213, Pennsylvania.}
\thanks{Praneetha Vaddamanu are with the Microsoft, 98052, Washington}}

\markboth{Journal of \LaTeX\ Class Files,~Vol.~14, No.~8, August~2021}%
{Shell \MakeLowercase{\textit{et al.}}: A Sample Article Using IEEEtran.cls for IEEE Journals}


\maketitle

\begin{abstract}
Recently remarkable progress has been made in 
synthesizing realistic human photos 
using text-to-image diffusion models. 
However, current approaches face degraded scenes, insufficient control, and suboptimal perceptual identity.
We introduce CustomEnhancer, a novel 
framework 
to augment 
existing 
identity customization 
models. 
CustomEnhancer is a zero-shot enhancement pipeline 
that leverages face swapping techniques,   pretrained  diffusion model,
to obtain additional representations  in a zero-shot manner for encoding into personalized models. 
Through our proposed triple-flow fused PerGeneration approach, which identifies and combines 
two compatible counter-directional 
latent spaces to manipulate a 
pivotal space of personalized model, we unify  the generation and reconstruction processes, realizing 
generation from three flows. 
Our pipeline also 
enables comprehensive training-free control over the generation process of personalized models,  offering  precise 
controlled personalization for them  
and eliminating the need for controller retraining for per-model.
Besides, to address the high time complexity of null-text inversion (NTI), we introduce ResInversion, 
a novel inversion method that performs noise rectification 
via 
a pre-diffusion mechanism, reducing the inversion time by 129×.
Experiments demonstrate that CustomEnhancer 
reach SOTA results at  scene diversity, identity fidelity, training-free controls,  
while  also showing  the efficiency of our ResInversion over NTI. The code 
will be made publicly available upon paper acceptance. 
\end{abstract}



\begin{figure*}[t]
\centering
\vspace*{-.1cm}
  \includegraphics[width=0.98\textwidth]{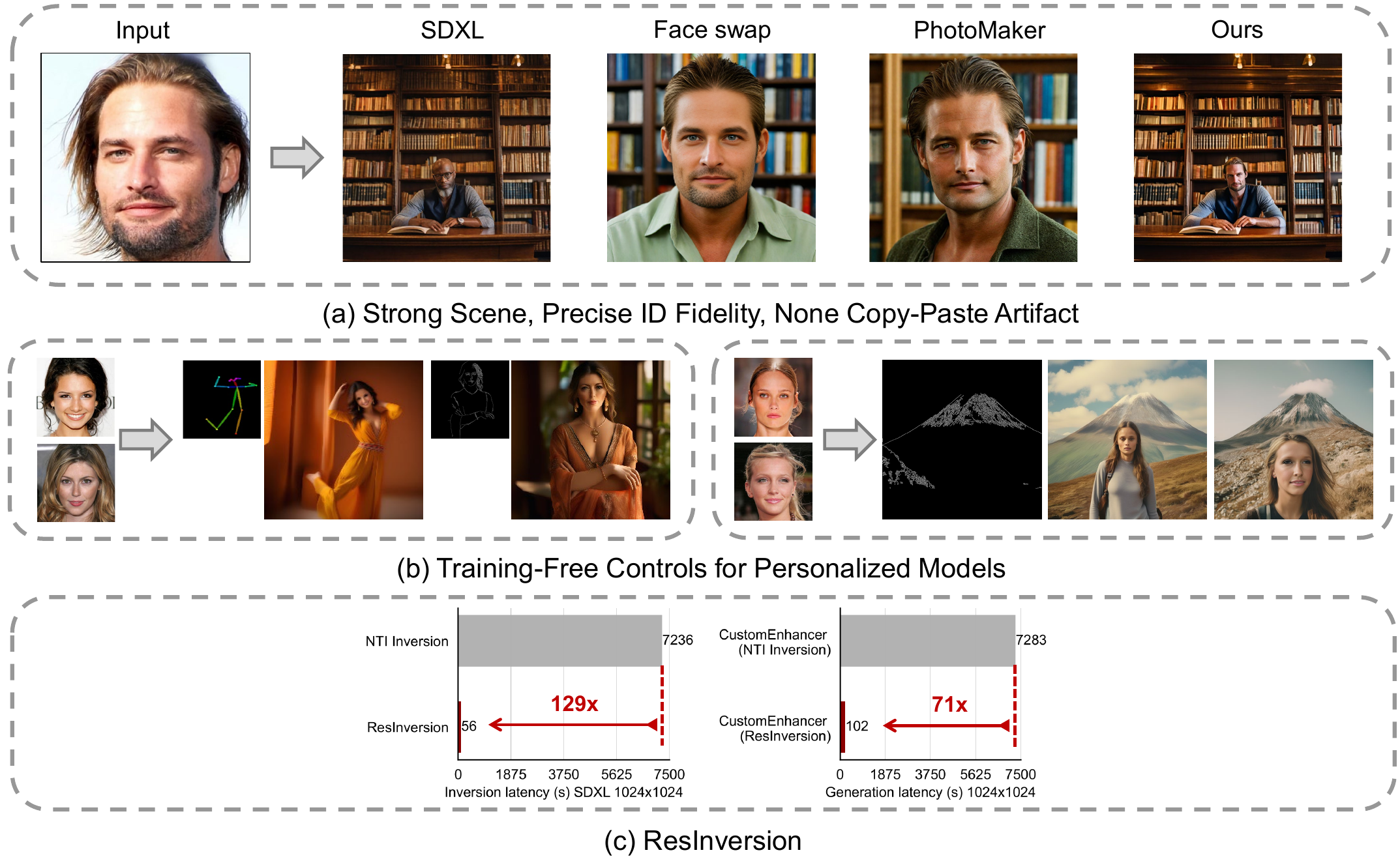} %
\caption{ Given a single image of input ID, our CustomEnhancer  (a) enables  SOTA identity customized 
  generation 
  with diverse large-scale scene, 
  further ID fidelity,  
 and zero artifacts of copy-paste, and (b)  enables comprehensive training-free controls for personalized models, offering 
 controlled personalization on human and both scene and eliminating inefficient per-model controller retraining, and
    (c)  Our ResInversion achieves 
    129× speedup over NTI inversion, and making our full pipeline  71× faster than NTI-based implementations.
    }
  \label{problem}
\end{figure*}

\section{Introduction}

In recent years, personalized  human photo synthesis, which aims to generate images consistent with the character's identity in reference images,  has made significant progress. 
Early methods, such as GANs~\cite{goodfellow2020generative}, were constrained by their generation capabilities and struggled to achieve satisfactory output. 
With the rise of large-scale text-to-image diffusion model~\cite{ramesh2022hierarchical, saharia2022photorealistic, rombach2022high}, personalized human photo Synthesis 
based on diffusion  models has seen great success. 

Current diffusion-based personalization (customization) methods   
can be categorized into two paradigms:
Test-time adaptation and training-based methods. 
Test-time adaptation 
methods such as DreamBooth~\cite{ruiz2023dreambooth}, Textual Inversion~\cite{gal2022image}, and Low-Rank Adaptation (LoRA)~\cite{hu2021lora} finetune pretrained models on input images during inference 
to enable identity-preserved generation. 
In contrast, 
training-based methods such as PhotoMaker~\cite{li2024photomaker}   and InstantID~\cite{wang2024instantid} 
adopt a more scalable finetune 
over   large-scale 
face-centric datasets during training phase to learn identity preservation capabilities in the model, eliminating test-time finetuning.   
These methods 
show 
faster inference speeds and better identity,  making them 
an emerging trend. 

However, we observe three major bottlenecks of current training-based personalization methods: 
I) Degraded scene generation 
capability: The extensive tuning on face-centered datasets 
causes existing personalized models 
to lose scene (context excluding the face) generation capabilities. They tend to generate only faces, abandoning backgrounds and bodies. 
II) Control module incompatibility:  The intensive  training isolates these models  from the control modules (ControlNet's pose/canny guidance) trained for the pretrained SDXL model, 
significantly reducing controllability and  necessitating inefficient control modules retraining for per-model. 
III) Suboptimal perceptual identity: Despite their specialization, these methods still exhibit an identity gap 
from reference images, due to current neural network-based 
extractor they rely on 
extracting mostly high-level features lacking concrete perceptual facial features.

Therefore, in this paper, we propose CustomEnhancer, a novel 
method offering 
diverse scenes, 
training-free controls, and 
identity fidelity. (see Fig. \ref{problem}). 
The key innovation of our method lies in:
I) A zero-shot enhancement pipeline: 
Our CustomEnhancer pipeline incorporates pretrained SDXL and face swap techniques to provide diverse scene representations and concrete perceptual facial representations in a zero-shot manner, respectively.  After fusing them into a shared 
image space, 
these representations are encoded into the personalized models’ generation process by our pipeline. 
II) Bidirectionally manipulated diffusion (BiMD) method: 
To integrate the generation of above  representations and  personalized model's customized identity representations, 
BiMD employs 
a dual-path latent manipulation paradigm 
that identifies and combines compatible counter-directional 
latent spaces: forward (generation) and backward (reconstruction) latent spaces. By 
intervening at a pivot flow of the personalized model through these two complementary spaces, we effectively unified 
the generation and reconstruction processes, 
endowing 
personalized models  
with additional information and capabilities. 
III) ResInversion: To address the high time complexity of inversion method of NTI, we introduce ResInversion, a novel inversion approach that performs noise rectification using the residual noise via 
a pre-diffusion mechanism, 
significantly reducing the time complexity of inversion by 129×. 
IIII) Training-free controls for personalized models: By integrating SDXL's control modules into our pipeline's SDXL component, 
we establish a training-free control framework that seamlessly embeds existing guidance to personalized models, enabling  control  over not only human subjects but also non-primary generation targets, the environmental elements, which remains unaddressed in prior work. 
It offers the controlled photorealistic personalization for these models and eliminates the computationally expensive control modules retraining. 

Our proposed CustomEnhancer 
is a plug-in method, compatible with any diffusion-based model (eg. PhotoMaker, InstantID, etc.). 
It  redistributes 
the scene and identity task to specialized auxiliary models, effectively solving the  copy-paste problem between  character and context. 
The main contributions of our paper are as follows:

\begin{enumerate}
    \item We propose CustomEnhancer pipeline capable of providing diverse scenes, training-free controls, and perceptual identity, 
    offering controlled personalization and eliminating inefficient controller retraining. 
    \item  We propose BiMD approach to integrate reconstruction into generation, 
    which was nonexistent in previous works. 
    \item  We propose Resinversion approach that enables fast inversion while maintaining reconstruction fidelity. 
    \item We realize NTI in more complex 
    SDXL framework, and implement in both  PhotoMaker and InstantID methods. We extend CustomEnhancer to other applications. 
    \item  Experiments demonstrate the superiority 
    of CustomEnhancer in 
    scene, control, and identity, without exhibiting copy-paste artifacts. 
\end{enumerate}

\section{Related Work}
\label{sec:formatting}

\subsection{Personalization in Diffusion Models.}
Traditional approaches, such as DreamBooth~\cite{ruiz2023dreambooth}, Textual Inversion~\cite{gal2022image}, Low-Rank Adaptation (LoRA)~\cite{hu2021lora}, and IP-Adapter~\cite{ye2023ip}  rely on extensive fine-tuning during the test phase to adapt the model to specific identities or concepts. These methods are time-consuming and require significant computational resources. Their identity preservation and image diversity is bad by finetuning on only few input images. 
Recent research has focused on single forward pass methods that bypass the need for fine-tuning, offering more efficient personalization. However, in methods like PhotoMaker~\cite{li2024photomaker} and InstantID~\cite{wang2024instantid}, the general generation ability has lost by the large-scale finetune, and the perceptual identity 
is weaker compare to reference image. 
Our work introduces an novel BiMD method integrating more representations to personalized model's generation process to overcome these limitations. 

\begin{figure*}
    \centering 
    \includegraphics[width=0.96\textwidth]{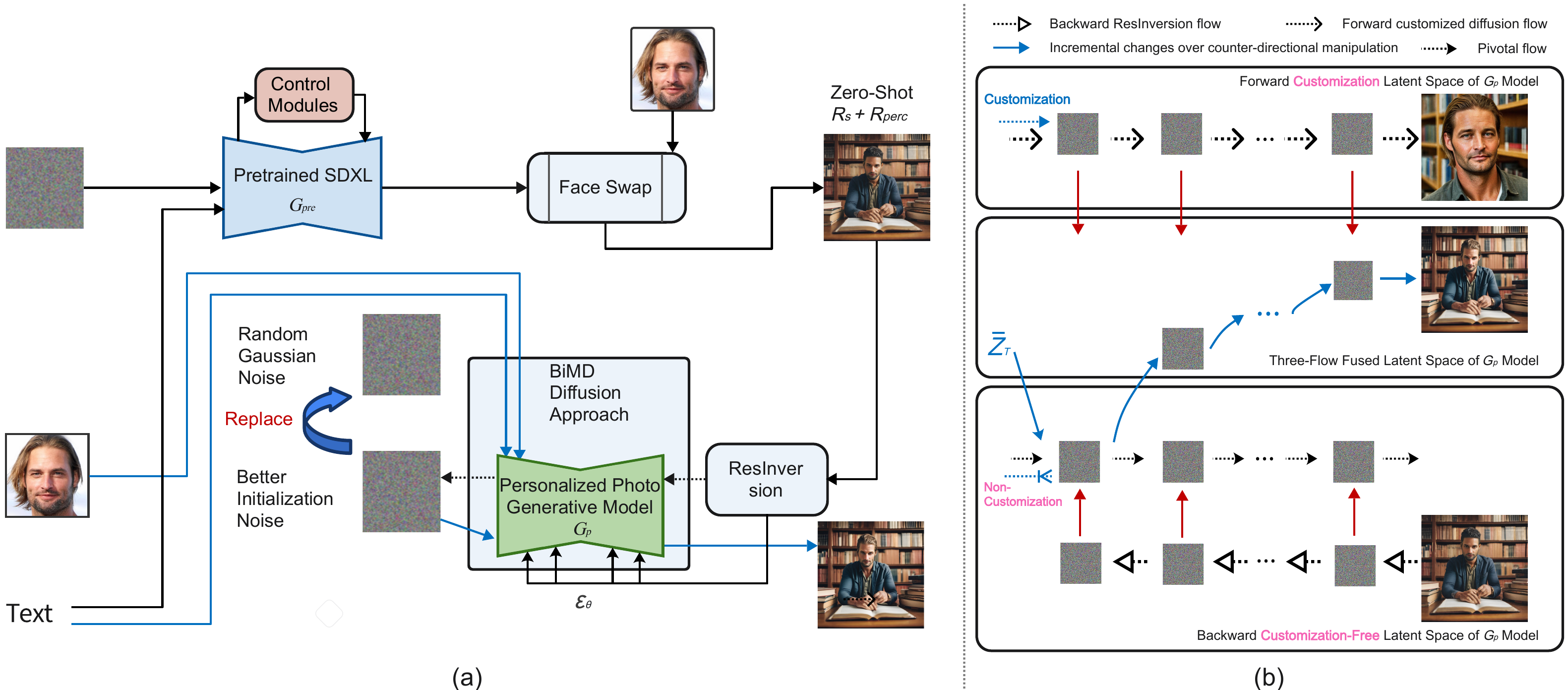} 
    \caption{
    Overview of the proposed: 
    (a) CustomEnhancer pipeline and (b) BiMD approach within personalized model. 
We first leverage pretrained SDXL and face swap techniques to provide diverse scene representations  and perceptual facial
representation in zero-shot manner. 
Then we use  BiMD to integrate the generation of these representation and $G_p$'s customized facial representation. 
BiMD employs compatible counter-directional latent spaces: forward customized generation (customization space) and backward 
ResInversion for reconstruction (customization-free space) latent spaces to intervene  
at a pivotal space, unifying the generation and reconstruction 
processes. 
Through incorporating SDXL's control modules upon SDXL component, training-free controls can be integrated into  personalized models' generation process through our pipeline.
    } \label{fig:pipeline}
\end{figure*}

\subsection{Inversion in Diffusion Models.}
Traditional Inversion methods, such as Denoising Diffusion Implicit Models (DDIM) inversion~\cite{dhariwal2021diffusion,song2020denoising}, are employed to reverse the diffusion process, enabling modifications to real images while retaining their original attributes. DDIM inversion works by reversing the ODE process in the limit of small steps, effectively running the diffusion process backward. However, in practice, a small error accumulates at each step.
Inversion of a given image in the context of diffusion models is an active area of research, and recent advancements, such as Null-text Inversion~\cite{mokady2023null} and ReNoise Inversion~\cite{garibi2024renoise}, have significantly improved upon the limitations of DDIM inversion. Null-text Inversion enhances the inversion process by 
optimizing the unconditional null-text embedding. 
This optimization helps to compensate for the inevitable reconstruction errors that occur when applying DDIM inversion directly, while preserving the editability of real images. ReNoise Inversion refines the inversion process, particularly for models with fewer denoising steps, 
enhancing reconstruction accuracy. 
It iteratively renoises the latent at each step, improving the predicted points along the forward diffusion path. 
By leveraging the inversion process, we build a path for information transferred to the personalized model's generation process.

\section{Method}

The goal of our method is to improve the  scene diversity, controllability, and identity fidelity for personalized models through purely generative means, without requiring additional environmental or identity images. 
The above abilities 
are mainly brought by our proposed zero-shot enhancement pipeline and BiMD approach, which innovatively 
combine 
information from both the forward and inverse  
diffusion processes, 
integrating  high-fidelity  reconstruction into generation.    
The  latency improvement is attributed to the proposed ResInversion that facilitates near real-time inversion for large-scale diffusion models. 
Our CustomEnhancer pipeline is illustrated in Fig.~\ref{fig:pipeline}. Below, we describe each component in detail.


\subsection{Zero-Shot diverse Scene  \& Perceptual facial features.}

Our approach 
builds upon the key observation that pretrained SDXL models, from which these personalized models are finetuned, inherently encode rich 
scene priors through their large-scale training on diverse datasets. Pretrained SDXL model also demonstrates exceptional capability in generating physically plausible human-context compositions. 
To leverage these capabilities, we propose to extract SDXL’s rich scene representations and embed them into personalized models' generation,  
enabling zero-shot scene encoding through  text prompt alone without requiring additional environmental references. 
Specifically, we employ SDXL to produce detailed scene images containing 
an identity-agnostic human character, guided by text prompt identical to 
the personalized models (Scene control is detailed in subsequent). 
The generated image can be represented as:
\begin{equation}
   I_P = R_{S} + ID_{ag} 
\end{equation}
where $I_P$ denotes the image generated by the pretrained model, $R_S$ represents the scene representation, $ID_{ag}$ denotes the identity-agnostic facial features. 

\begin{figure}
    \centering 
    \includegraphics[width=0.48\textwidth]{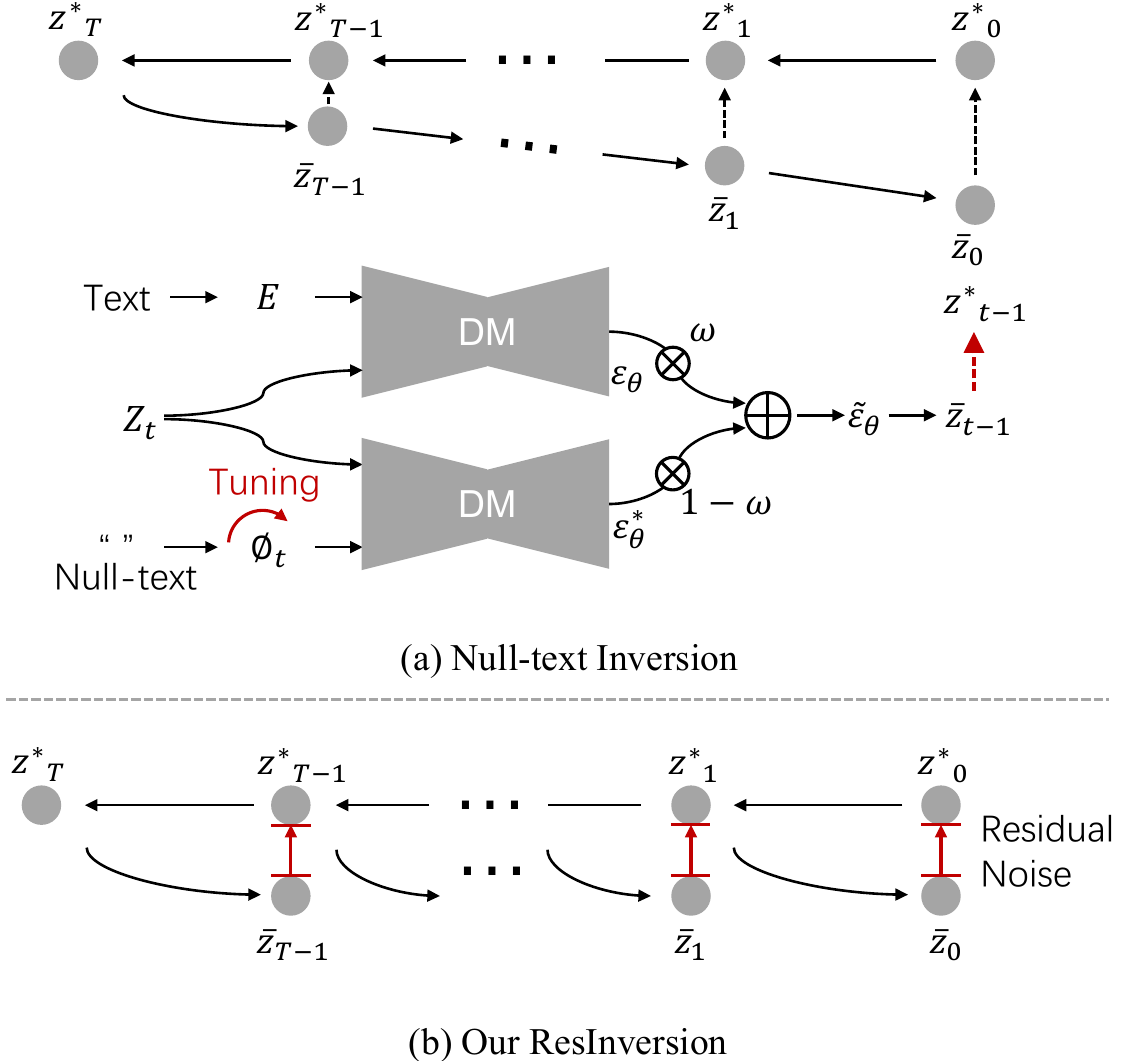}
    \caption{ (a) Null-text inversion. (b) ResInversion.  
} \label{nulltex}
\end{figure}

Personalized models' neural network-based facial feature extractor primarily 
capture high-level features while lacking sufficient concrete 
perceptual features. 
We propose to adopt 
face swapping (FS) technique that additionally supplies concrete perceptual facial 
representation 
(eg. precise geometric shapes, accurate landmarks, and fine-scale attributes) for personalized models. Specifically, we employ the FS  to inject reference identity into the above identity-agnostic image through  anchor-point alignment. 
Then, the image now can be represented as:
\begin{equation}
   I_P = R_{S} + R_{perc} 
\end{equation}
where $R_{perc}$ denotes concrete perceptual facial representation obtained from FS. 
The limitations of FS lie in 
modifying 
only the facial subjects instead of other attributes like hairstyle and skin color,  and introducing misaligned edges and blurring in face regions, 
which can be effectively resolved by our subsequent step. 
Next, we encode these representations into the generation process of the personalized model.
Notably,  FS integration enables our pipeline to handle identity preserving in small localized regions and viewpoint changes, a persistent challenge  for generative methods.

\subsection{ResInversion.}
We first describe NTI and then our ResInversion. 
NTI 
is an inversion method that achieves  accurate inversion of real images while preserving the editing ability. 
As is depicted in Fig. \ref{nulltex} (a), 
it begains by generating initial noise trajectories, $Z_0^*, \cdots, Z_T^*$, using DDIM inversion serving as an optimization anchor. To address the accumulated error in DDIM inversion, in which each step deviates the noise trajectory from its expected path, 
NTI utilizes a  
tuning mechanism named Null-text Optimization. 
NTI optimizes the unconditional text embedding 
to adjust the unconditional denoising part for solving the accumulated error and achieving high-fidelity reconstruction. 
NTI was initially proposed in smaller architectures (SD1.5 model), and we realize it in the more challenging SDXL framework which  serves as the  backbone of current SOTA personalized models. 

However, tuning text-embedding in NTI requires  propagating  through the full diffusion model every optimization step (Fig. \ref{nulltex} (a)). 
Although it 
remains computationally tractable for compact architectures like SD v1.5,  
the text-embedding optimization becomes 
prohibitive 
time complexity in larger models. 


We propose ResInversion, a novel inversion approach that directly identifies and compensates for noise deficiency in each diffusion step to address trajectory deviation due to the accumulated error. 
Specifically, as illustrated in Fig. \ref{nulltex} (b),  1) 
We 
first employ DDIM inversion to obtain initial noise trajectories, $Z_0^*, \cdots, Z_T^*$ 
as the anchor. 
2) 
Then, in our ResInversion, we propose to  explicitly records residual noise through a pre-diffusion process. 
During this preliminary diffusion, we performs stepwise noise correction and 
simultaneously documents 
the residual noise $\{\epsilon_{r,0}, \cdots, \epsilon_{r,t}, \cdots, \epsilon_{r,T}\}$, denoted as
\begin{equation}
     \mathlarger{\mathlarger{\epsilon}}_{r,t} =  \mathlarger{\mathlarger{\epsilon}}_{DDIM, t} -  \mathlarger{\mathlarger{\epsilon}}_{pre\text{-}d, t} 
\end{equation}
where $\mathlarger{\mathlarger{\epsilon}}_{r,t}$ denotes the documented residual noise component at time step t, $\mathlarger{\mathlarger{\epsilon}}_{DDIM,t}$ 
represents the removed noise derived from DDIM inversion trajectory between timestep $t$ and $t-1$,
$\mathlarger{\mathlarger{\epsilon}}_{pre\text{-}d,t}$ indicate the denoised part 
by pre-diffusion process of the model with the text prompt.  
These residuals then rectify noise discrepancies in forward step, aligning trajectories with pivot flow targets
$Z_0^*, \cdots, Z_T^*$. 
ResInversion eliminates the need for iterative optimization, achieving significant 
speedup inversion 
while maintaining reconstruction fidelity.

\subsection{BiMD approach.}
To integrate 
the representations from  $I_P$'s image space into personalized model $G_p$'s generation process and achieve integrated generation, 
we introduce BiMD approach.  As  illustrated in Fig. \ref{fig:pipeline} (b), BiMD diffusion method consists of three synergistic flows in $G_p$'s denoising process: 
1) Pivot flow: We 
utilize DDIM inversion to obtain initial noise trajectories as pivot flow. 
2) Backward ResInversion flow:  We 
invert the image $I_P$ into the  $G_p$ by ResInversion based on the pivot anchor, projecting the representations in $I_P$ 
from the image space into the dispersed 
residual noise space. 
Different from NTI-like inversion method inverting the image into the fully functional $G_p$, 
we propose to 
invert image into the partial functional space, customization-free space where $G_p$ generates non-specific identity in process, 
This aims to 
preserve  $G_p$'s personalized capabilities from been overwritten by backward flow. 
Notably, the backward ResInversion flow is 
compatible with the pivot flow due to ResInversion’s inherent mechanism, 
which enables pivot flow to faithfully 
reconstruct image $I_P$ without information loss. 
3) Forward customization flow: We 
integrate the 
customized denoising process 
from $G_p$ 
into the pivot flow,  which  incorporates $G_p$'s identity customization capabilities for the current latent. 
%
This 
customized generation 
flow 
remains compatible with the pivot flow 
as it can customize  arbitrary free generation 
process of $G_p$ from customization-free space, of which the pivot flow is one instance. 
Through identifying these two compatible spaces and establishing a  dual-path 
latent manipulation from counter-direction for a pivot flow, BiMD achieves  integration  
of the customized identity and the backward representations 
within a unified generation process, which able to transfer information from both forward $G_p$'s customization and backward ResInversion's reconstruction. 
In sampling procedure of diffusion process, the updating formula of BiMD is: 
\begin{equation}
\begin{split}
Z_{t-1}=&\frac{1}{\sqrt{\alpha_t}}\bigg\{Z_t-\frac{1-\alpha_t}{\sqrt{1-\bar{\alpha}_t}} \Big\{ \mathlarger{\mathlarger{\epsilon}}_\theta(Z_t,t) \\
&+  \mathlarger{\mathlarger{\epsilon}}_{Res}(Z_t,t) - \mathlarger{\mathlarger{\epsilon}}_{G_{p, cus}}(Z_t,t) \Big\} \bigg\}
\end{split}
\end{equation}
where $\alpha_t$ denotes noise scheduling parameters, $\epsilon_\theta$ represents network predicting noise, 
$\mathlarger{\mathlarger{\epsilon}}_{Res}$ 
indicates the residual noise  of backward ResInversion at time step $t$ under customization-free spaces, 
$\mathlarger{\mathlarger{\epsilon}}_{G_{p, cus}}$  
represents predicted noise of forward $G_p$ 
under identity customization space. 
%
%
Through iterative refinement via model's diffusion process, we generate final images that preserve bidirectional representations, maintaining both reconstruction representation and customized 
attributes.

The key distinction between our approach and other DDIM inversion-based methods  lies in its information supplementation capability  for the inverted model. Conventional DDIM inversion fails to retain information during the regeneration process~\cite{mokady2023null}, as it predominantly relies on the inverted model's generative capabilities to recreate whole content,  representing a reuse of the model's existing generative capacity. Our approach  supplements the model with additional information and abilities in their diffusion process that the original model lacked. 
Furthermore, although BiMD has three flows, it employs the diffusion processes under a single model via a forward and inversion paradigm, as opposed to blending of diffusion processes from multiple models, thereby avoiding 
artifacts arising from domain gaps of various models, 
such as inconsistent copy-paste effects.

\subsection{Training-free controls.}
Proposed pipeline establishes a training-free control paradigm for diffusion-based personalized models, 
through integrating 
pretrained SDXL's control modules into 
SDXL component in our pipeline, 
as depicted in Fig. \ref{fig:pipeline} (a) upper branch.   
Our pipeline seamlessly embeds 
the control into the generation process of personalized models, 
enabling comprehensive training-free controllability across personalized systems. The control embedding can be 
denoted as: 

\begin{equation}
I_P = R_{S} + R_{perc} + C_a
\end{equation}

where $C_a$ indicates arbitrary control modules trained for SDXL. 
This has two key advantages:  
(1) Enabling controlled photorealistic personalization on both primary generation targets, i.e., the human subject, and non-primary generation targets, such as environmental elements. 
%
Specifically, precisely controlling non-primary generation elements presents a significant technical challenge, a capability notably absent in prior work due to the inherent difficulty of control  non-primary generation elements while maintaining the quality of primary subjects. 
(2) Eliminating the  computationally expensive retraining of control modules  typically required for personalized models.

\section{Experiments}

\begin{figure*}[!ht]
    \centering
    \includegraphics[width=0.94\textwidth]{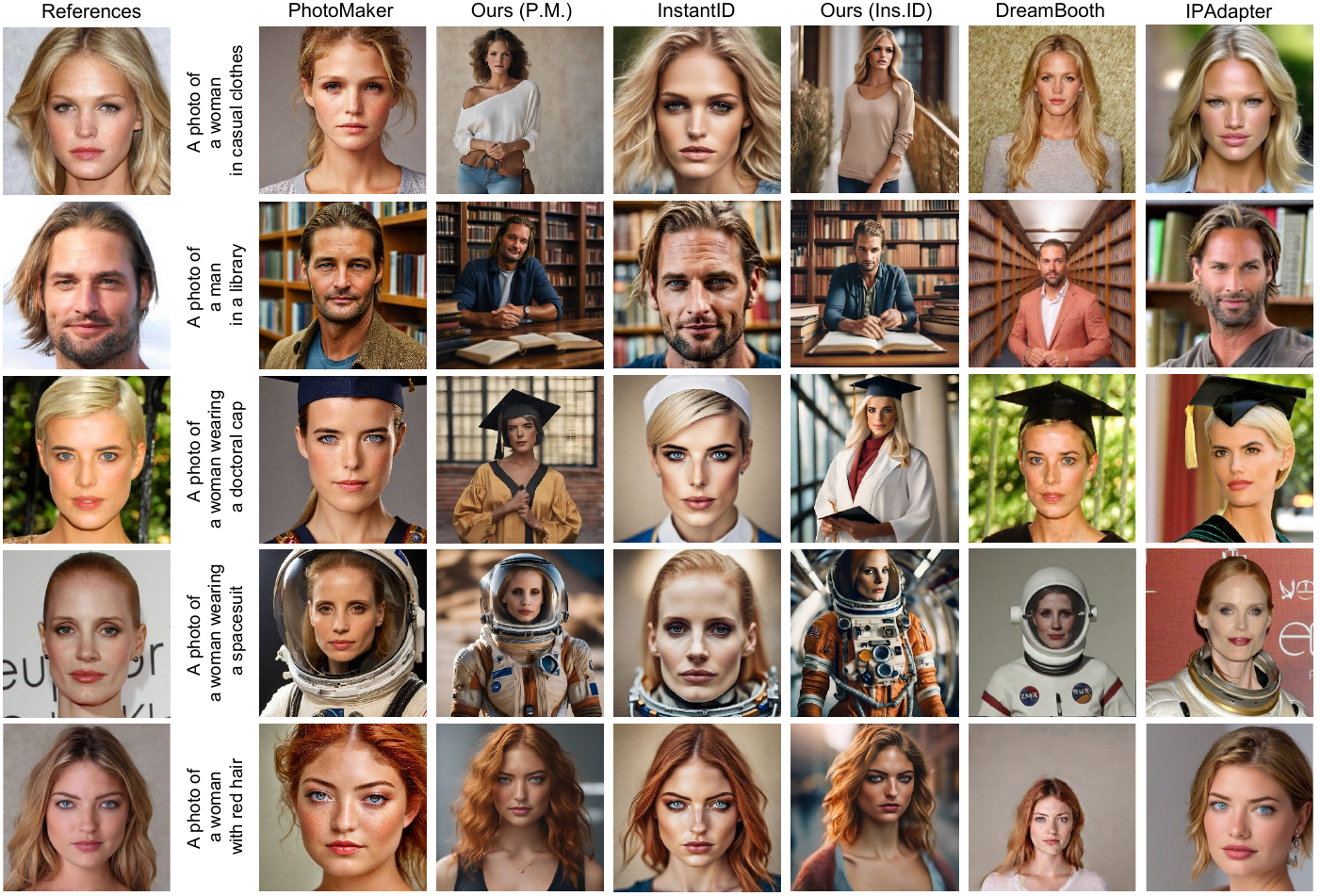} 
    \caption{
Qualitative comparison on identity-preserved 
recontextualization samples. 
We compare our method with the SOTA PhotoMaker, InstantID, Dreambooth, and IPAdapter under various 
identities  and prompts. Our method is plugged into PhotoMaker and InstantID, denoted as  ours (P.M.) and ours (Ins.ID).  We observe that our method significantly enhances the scene diversity, identity fidelity, and produces  high-quality images without copy-paste artifacts. 
    } \label{figresults}
\end{figure*}

\begin{table*}[!ht]
    \tablestyle{5pt}{1.15}
    \centering
    \caption{Quantitative 
    results on PhotoMaker and InstantID and our method when they plug in.} 
    \resizebox{0.98\linewidth}{!}{
    \begin{tabular}{lccccccc} 
    \toprule
                  &Scene Div.↑ (\%) &Face Sim.↑(\%)   & CLIP-T↑ (\%) & DINO ↑ (\%) & CLIP-I↑ (\%) & CLIP-IQA↑ (\%)  & FID↓    \\ \midrule
        CustomEnhancer (P.M.) & \textbf{60.0} & \textbf{67.3}  & \textbf{31.7} & \textbf{85.4} & \textbf{82.4}  & \textbf{89.6} &\textbf{416.3}  \\ 
        PhotoMaker & 52.2 & 55.7   & 27.8 & 84.9 & 80.8 & 87.4 &473.9 \\ 
        
        CustomEnhancer (Ins.ID) & \textbf{58.5} & \textbf{70.5}   & \textbf{32.4} & \textbf{85.6} & \textbf{78.2} & \textbf{88.7} & \textbf{423.6} \\ 
        InstantID & 52.8 & 61.1  & 29.6 & 84.7 & 76.3 & 86.4 & 496.1 \\ 
        \midrule
        DreamBooth & 54.6 & 45.2  & 29.2 & 73.9 & 68.9 & 84.7 & 534.2 \\ 
        IPAdapter & 55.4 & 52.6  & 26.5 & 79.2 & 78.6 & 86.9 & 483.8 \\ 
        \bottomrule
    \end{tabular}
    }
    \label{fs}
\end{table*}

\subsection{Experimental Setup}  

\paragraph{Evaluation dataset}
We used the CelebA-HQ dataset~\cite{karras2017progressive} as our base dataset. We select the IDs from CelebA-HQ with more than 20 images to form our dataset. 
In total, we identified 30 IDs from CelebA-HQ to comprise our dataset,  
which has a similar ID number 
as~\cite{li2024photomaker}. 
We  prepare 40 prompts for each ID, which has a variety of attributes and backgrounds following~\cite{li2024photomaker}. 

\paragraph{Evaluation metrics}
Following
~\cite{li2024photomaker} we use face similarity, DINO ~\cite{caron2021emerging} and CLIP-I  ~\cite{gal2022image} metrics to measure the ID fidelity. For the face similarity, 
we use is MTCNN~\cite{zhang2016joint} as the detection model and the face embedding  is extracted by 
FaceNet 
~\cite{schroff2015facenet}   
following ~\cite{li2024photomaker}. 
For each ID, we have 20+ images as mentioned above. In the evaluation, 
we randomly  select one showing the full face and the other images of this ID are used for evaluating the face similarity by averaging.  
To evaluate the quality of the generation, we use the FID metric~\cite{heusel2017gans} and CLIP-IQA~\cite{wang2023exploring}. We calculate the FID between generated image and HumanArt dataset~\cite{ju2023human} containing high-resolution images of faces,   scenes, and interactions.    We also adopt CLIP-T~\cite{radford2021learning}  metric to measure the prompt fidelity and the FID metric to evaluate the quality of the generation.  
For the scene diversity, we propose a metric, termed Scene 
Diversity, to measure the diversity of the regions excluding  facial region, which
detects and crops out the face region and calculates the LPIPS ~\cite{zhang2018unreasonable} scores of the remaining images. 

\paragraph{Comparison methods} 
We select the SOTA  personalized photo generation method PhotoMaker~\cite{li2024photomaker} and InstantID~\cite{wang2024instantid} as the comparison methods, which are also the methods our pipeline plugs 
into. We also include DreamBooth~\cite{ruiz2023dreambooth} and IP-Adapter~\cite{ye2023ip} as benchmark methods. 
The experimental results are systematically compared against these approaches.

\paragraph{Implementation details}
To generate the scene and ID-agnostic image in our pipeline, we employ SDXL model \cite{podell2023sdxl} \textit{stable-diffusion-xl-base-1.0}. We adopt InsightFace Swapper \footnote{https://github.com/haofanwang/inswapper} as the face swapping model. 
The 50-step  DDIM sampler is employed for all diffusion models. All the experiments are conducted on 8 NVIDIA RTX A4500 GPUs. 


\subsection{Comparesion results} 

\paragraph{Scene diversity \& ID fidelity}
We conduct
both quantitative and qualitative comparisons of the generated
results of compared methods. For each personalized  method our pipeline plugs into, we prioritize using the official 
codes following their papers. 
The experiment  results for the quantitative comparisons
are presented in Tab.  \ref{fs}. The highest performance is shown in bold. The results show that our
method  generates  images with diverse scenes (Scene Div., FID) and higher ID fidelity 
 (Face Sim.), while ensuring overall 
 image quality (CLIP-IQA, FID). 
The qualitative comparisons are illustrated 
in Fig. \ref{figresults}.
The results demonstrate 
that our method produces 
images with high-quality scenes (complex background and detailed body feature), 
physically plausible human-context interaction 
(hands in the desk with book) without artifact of copy-paste, and enhanced facial regions with better ID fidelity. 
The overall image is generated via a unified 
diffusion process within personalized models, which  guarantees the image quality.
%

\paragraph{Large-scale scene by scene-invoking prompts}\label{scene-invoke}

We incorporate specific keywords into  the prompt and conduct experiments  to evaluate the performance of CustomEnhancer and personalized methods in large-scale scene generation tasks. 
\begin{figure}[!ht]
  \centering  \includegraphics[width=0.48\textwidth]{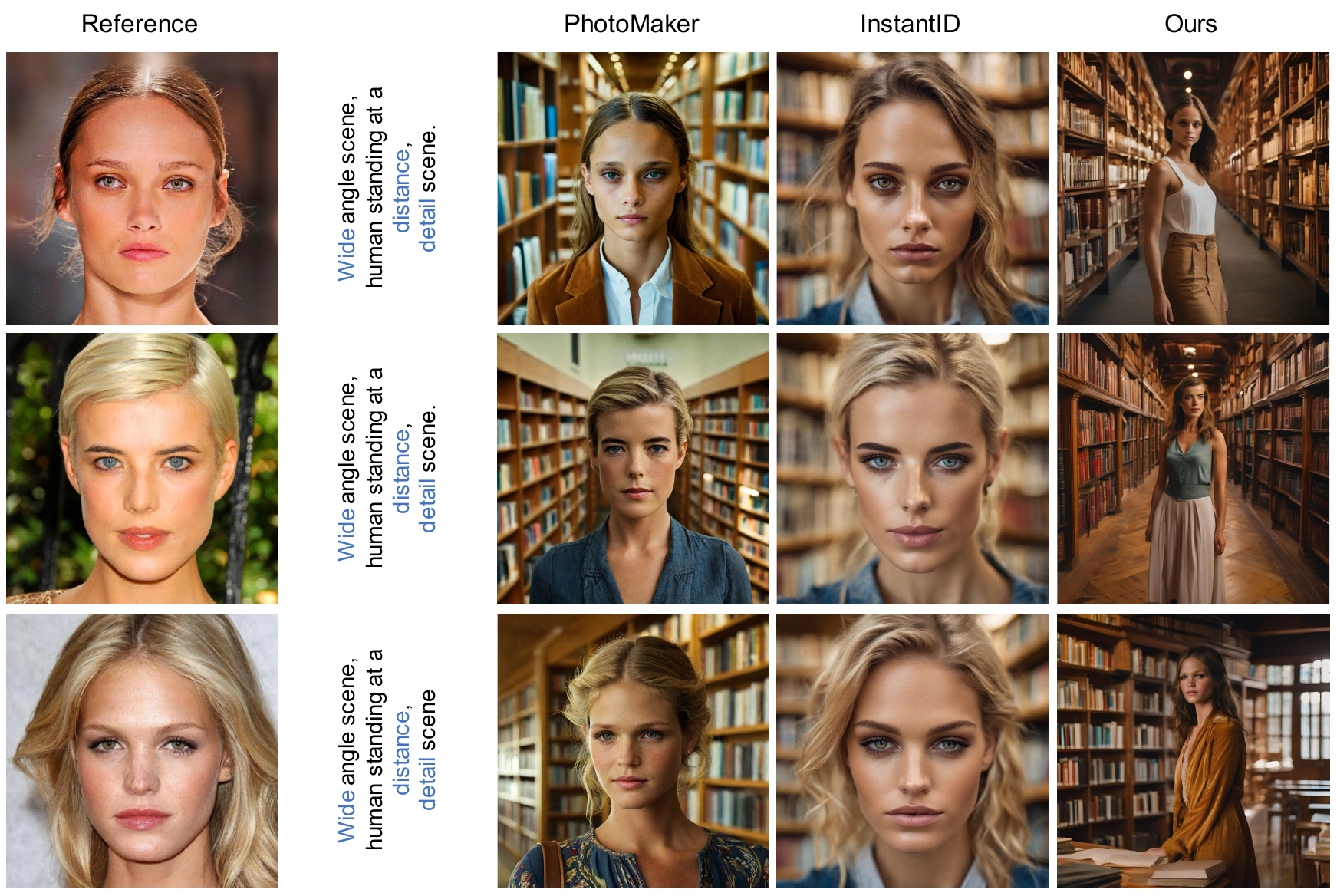}
  \caption{
  Comparison under scene-invoking prompts between PhotoMaker, InstantID, and our method. 
  Our method generated expansive and detailed scenes, high-quality details of the human, 
  reasonable character-context interactions, while personalized models fail. 
  }
  \label{wide}
\end{figure}
We included phrases ``wide angle scene", ``human standing in the distance", and ``detailed scene" into the prompts, ``A photo of a \texttt{<class word>} in a library", to generate large-scale 
scenes. The results are illustrated in Fig.  
Results show that although these personalized generation models produced more detailed scenes compared to those without additional prompts (see Fig. \ref{figresults}), they still generated only a small portion of the scenes, which are  entirely not satisfactory. In contrast, our method generated expansive and detailed scenes (Fig. \ref{wide} third column), while also producing high-quality details of the human character and their plausible 
interactions with the environment (hand on the table). 
By incorporating  scene-invoking texts 
in the prompts, our pipeline can further exert the scene generation ability in the pretrained SDXL 
and produce better scene  representations. 
In contrast, these personalized models have diminished 
scene generation ability. 
This demonstrates that, under the same prompts, our method consistently produces superior scenes, character details, and reasonable character-context interactions, with zero copy-paste artifact, compared to existing personalized models. 

According to the limitations discussed in PhotoMaker manuscript~\cite{li2024photomaker}, 
PhotoMaker  excels at generating half-length portraits, but performs 
poorly in generating full-length portraits. 
It is  a 
common challenge for current diffusion-based personalized models. By plugging our method into these personalized 
models, they can be enhanced with the capabilities of  generating full-length human  and diverse scenes. More importantly, our method also enhances them with 
training-free controls and ID fidelity.

\paragraph{More dataset results}

We conducted expanded experiments on the CFB dataset to 
evaluate our method's generalization ability. 
In CFB dataset, we select a frontal-facing image as input within images that represent  the same ID and utilize the remaining images of this ID for face similarity evaluation by averaging. 
The other metrics  are measured in the same way as the  main paper. Experimental results are presented in Tab. \ref{CFB}. 
The results consistently demonstrate proposed method's superior performance, confirming its effectiveness in enhancing personalized models with both scene diversity and ID fidelity. 
The robust performance across diverse datasets  validates the generalization and robustness of our proposed method.

\begin{table}[!ht]
\centering
\caption{Results of scene diversity and face similarity on CFB dataset.}
\begin{tabular}{lcccc} 
\toprule
Method &  Scene  Div.↑ & Face Sim.↑ &  CLIP-I↑ &  FID↓ \\ 
\midrule
        Our pipeline (P.M.) & 60.4 & 66.2  & 82.5 & 417.8 \\ 
        PhotoMaker & 52.6 & 54.4  & 81.0 & 477.3 \\ 
        Our pipeline (Ins.ID) & 58.9 & 69.6  & 78.4 & 428.6 \\ 
        InstantID & 53.0 & 60.3  & 76.8 & 498.5 \\ 
\bottomrule
\end{tabular}
\label{CFB}
\end{table}

\begin{figure*}[htbp]
    \centering
    \subfloat[Human controlled personalization.]{
        \includegraphics[width=0.90\textwidth]{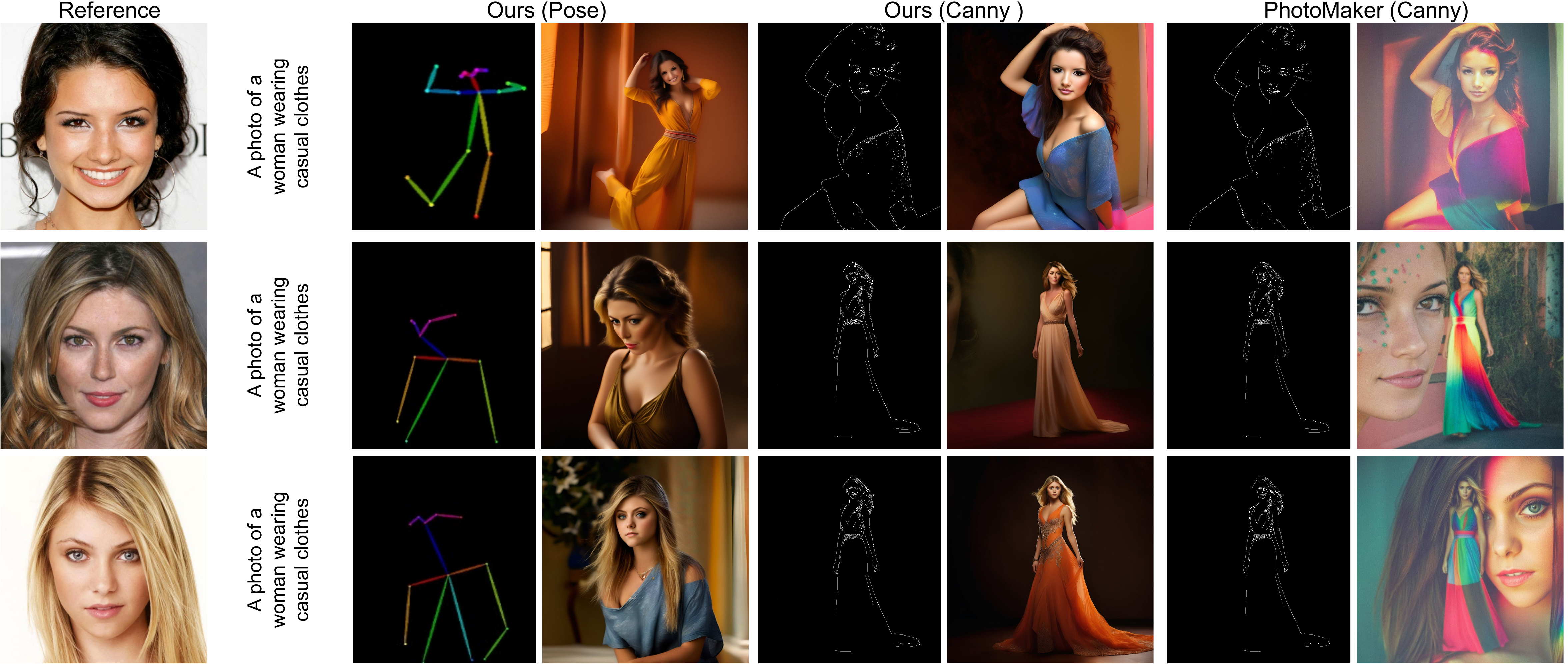} 
    }
    \hfill
    \subfloat[Environmental-elements controlled personalization.]{
        \includegraphics[width=0.90\textwidth]{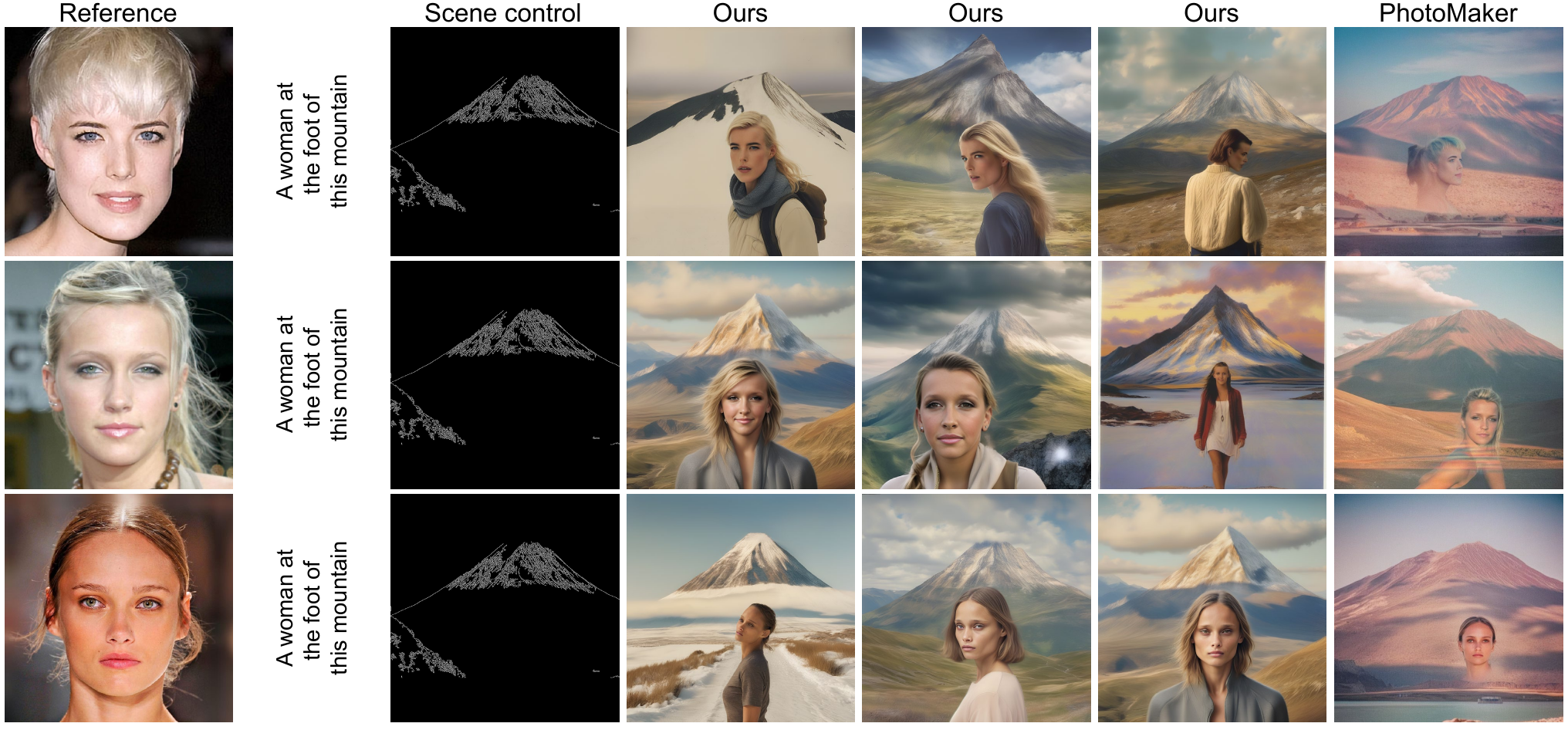}
    }
    \caption{  
    Training-free control for (a)  human controlled personalization, taking the openpose  and canny controls as examples,  
    and (b) environmental-elements controlled personalization, where our CustomEnhancer  enables controlled personalization on non-primary generation targets. elements.
    } \label{controlnet}
\end{figure*}

\paragraph{Training-free control: Human subjects and environmental elements controlled personalization}
We incorporate  OpenPose and Canny controlnets as representative examples in the SDXL component (Fig. \ref{fig:pipeline} upper branch) and conduct the human-controlled experiments in 
personalized models.  
For a fair comparison, we apply  the same controlnet module 
and prompt across all methods. 
As illustrated in Fig. \ref{controlnet} (a), 
the results exhibit that our method precisely adheres to control conditions while maintaining high image quality and identity fidelity, with no copy-paste artifacts.
In contrast, personalized models like PhotoMaker show incompatibility with these control modules, leading to corrupted image quality, poor identity preservation, and inconsistent outputs.
Our method does not interfere  with the identity diffusion process of these personalized models, ensuring  high image quality and artifact-free synthesis.   
Results generalize  effectively to other SDXL-compatible control modules, providing personalized models with training-free controls and removing the inefficient controller retraining.

We employ an  canny-based scene control in pretrained SDXL as example to conduct experiments on scene control. We use the identity prompt !A woman at the foot of this mountain' to generate the image for all compared methods. 
As shown in Fig. \ref{controlnet} (b), our method can obtain the marvelous scene (mountain), while retaining the high ID fidelity on the generated images. Environment and human subjects and controlled environment are composed naturally without copy-paste problem. Our method also enables  personalized methods with full-body generation capability, as well as detailed face generation on small regions of an image. In contrast, the existing personalized models not only generate poor image quality, but also struggle to condition  on the non-primary elements, generating basically no human in the image. 

\paragraph{ResInversion}


We implement the NTI-based CustomEnhancer in PhotoMaker and InstantID frameworks, while realizing NTI for inversion testing in the general SDXL model, followed by comprehensive comparisons with our ResInversion-based implementations.
Generation latency is measured with batch size=1 on a single GPU. The quantitative results are detailed in Table \ref{resinversion}. Our ResInversion makes our full pipeline 71× faster than NTI-based implementations in PhotoMaker and 65× in InstantID, reducing latency from hours to minutes. 
%
Although DDIM inversion achieves the fastest processing speed, it fails to accurately reconstruct the input image and merely utilizes the model's generative capabilities to produce new samples,  providing no additional accurate  information to enhance the model. 
In contrast, our ResInversion introduces only a small latency overhead while providing significantly distinguished reconstruction quality than DDIM inversion.

\begin{table}[!ht]
\centering
\caption{Generation latency by different inversion methods.}
\resizebox{1\linewidth}{!}{
\begin{tabular}{lcccc} 
\toprule
                 & PhotoMaker  (s) & InstantID  (s)    & SDXL (s)   \\
\midrule
NTI    &   7283  & 9237     & 7567        \\
ResInversion  &  102    & 141     & 98        \\
DDIM Inversion  &  54    & 82      & 52      \\
\bottomrule
\end{tabular}
}
\label{resinversion}
\end{table}

\begin{figure}
    \centering 
    \includegraphics[width=0.45\textwidth]{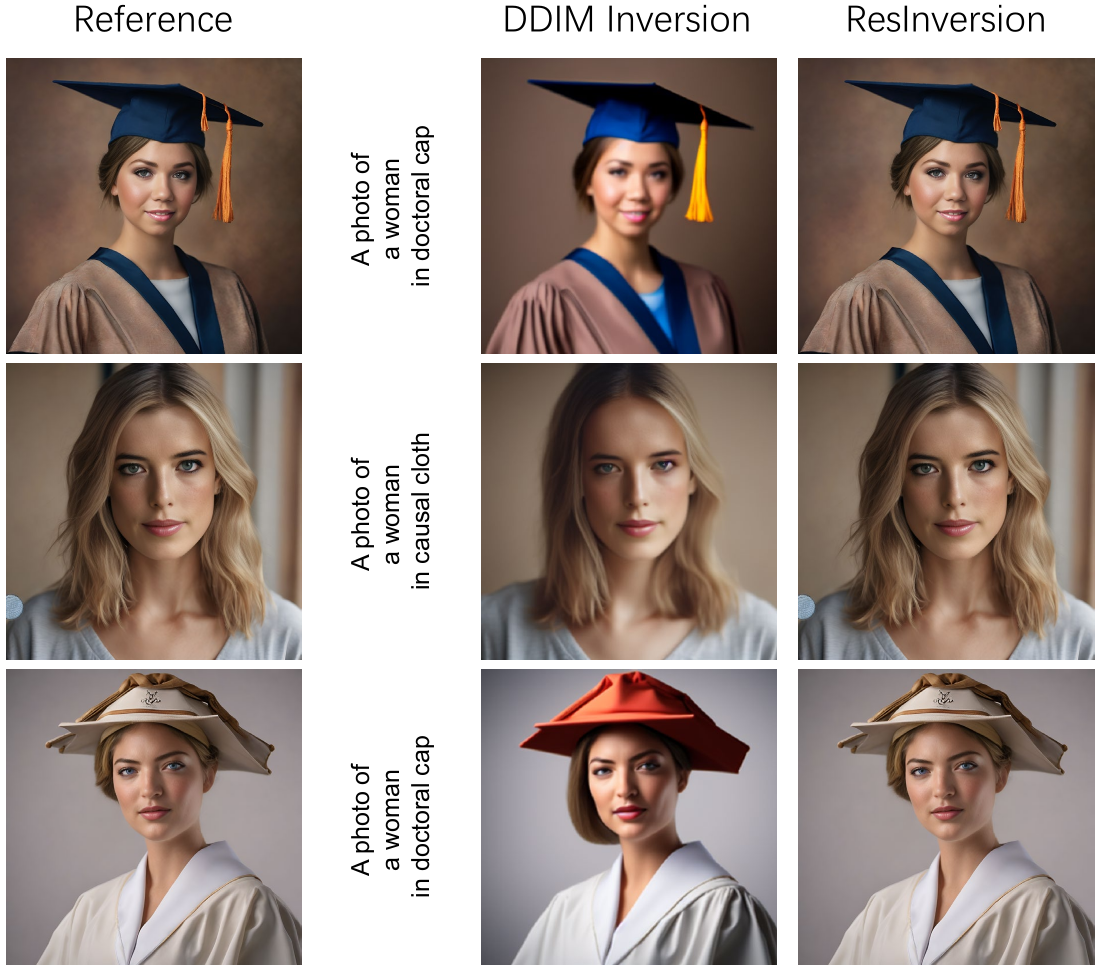}
    \caption{
    Qualitative results of ResInversion on reconstruction quality.   
    } \label{ablation1}
\end{figure}

\subsection{Ablation study}

\paragraph{Module ablation} 

We ablate each component in reverse order, presenting results after final output, FS 
component, SDXL component of our pipeline, and we also ablate the forward flow (AB. FWD flow) and backward flow (AB. BKWD flow) in BiMD. 
The qualitative results are illustrated in Fig.~\ref{ablation1}, which shows  that BiMD further modifies  the smectic area of human from the base image,  such as the hairstyle, skin color of face and arm, and the face identity,  while keeping  the background consistent (After Final and After FS). 
FS effectively injects the perceptual identity into the pretrained image (After FS and After SDXL), 
and backward flow faithfully 
reconstructs the inverse image (AB. FWD and After FS) and forward flow is able to customize the face (AB. BKWD).  

\begin{figure*}
    \centering 
    \includegraphics[width=0.9\textwidth]{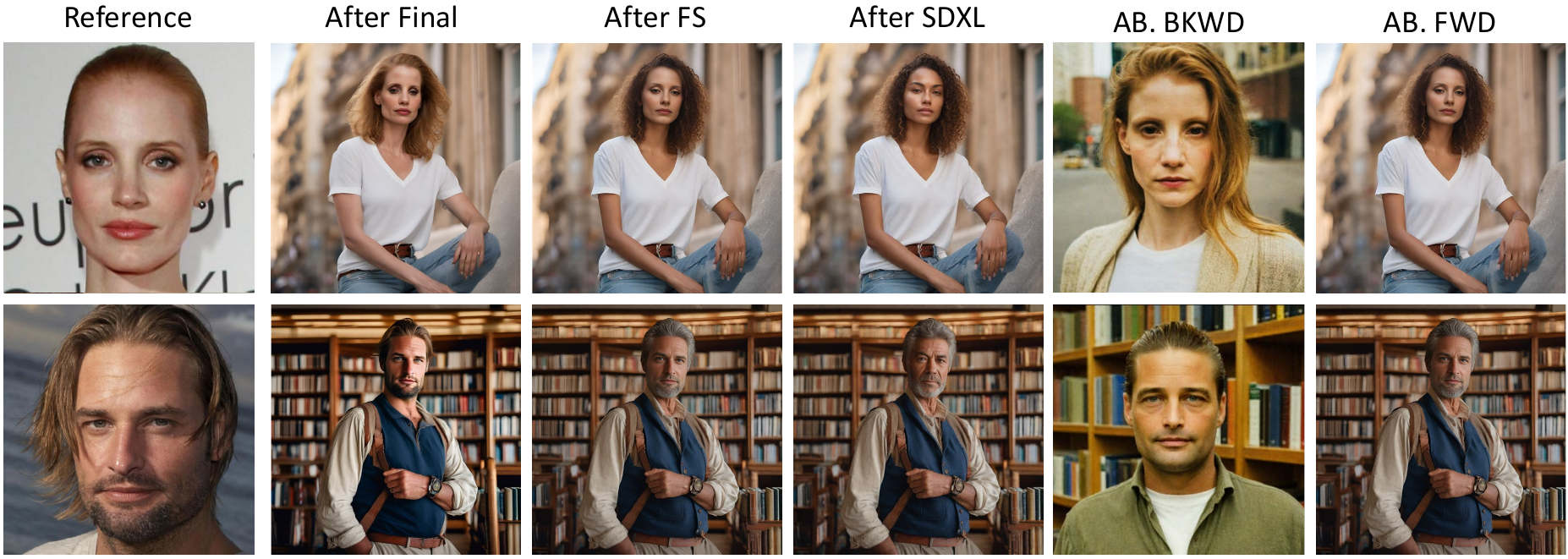}
    \caption{
    Ablation on components. 
    Results after final output, face swapping, SDXL components, and after ablating the FWD  and BKWD flow in BiMD. 
    } \label{ablation1}
\end{figure*}

The quantitative  results are summarized in Table. \ref{ablation}. (For the results of CustomEnhancer 
and personalized 
models, refer to 
Table. \ref{fs}). 
Results show that the face similarity in the output image of CustomEnhancer  is better than that of both   face swap and personalized models. This  demonstrates that the concrete perceptual 
identity, additionally injected through face swap, can further complete the 
identity information for neural  network-based extractors in personalized models. 


\begin{table}[!ht]
\centering
\caption{Quantitative Ablation results on our pipeline.}
\resizebox{0.98\linewidth}{!}{
\begin{tabular}{lccccc} 
\toprule
\multicolumn{5}{c}{CustomEnhancer (PhotoMaker)} \\
\midrule
                 & Scene DIV ↑ (\%) & Face Sim. ↑  (\%)     &CLIP-IQA↑ (\%) & FID↓  \\
Ab. BKWD      &   50.6  & 55.1               & 84.2      & 485.9 \\  
Ab. FWD flow &   58.2  & 63.5               & 84.5      & 428.2 \\
After FS    &   58.2  & 63.5               & 84.8      & 423.8        \\
After SDXL  &   58.2  & 18.7                & 89.5     & 419.7\\
Input image  &  33.8    & 77.4             & 82.3      & 441.9          \\ 
\bottomrule
\end{tabular}
}
\label{ablation}
\end{table}

\paragraph{Manipulation starting step.} 

\begin{figure}
    \centering 
    \includegraphics[width=0.48\textwidth]{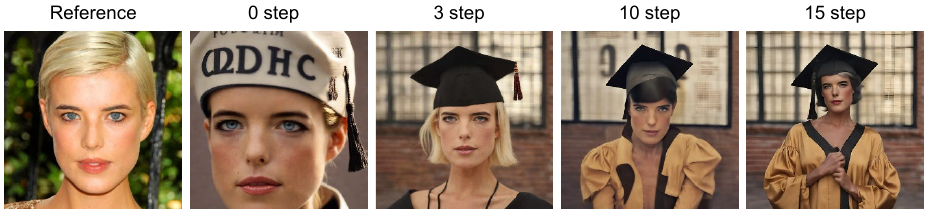}
    \caption{
    Ablation through MSS 
    step. 
    Smaller steps tend to generate  by PhotoMaker whereas larger steps by our pipeline.
    } \label{qulitysteps} 
\end{figure}

We ablate the manipulation starting step (MSS) of the forward and backward flow at the pivot flow within the 50-step DDIM sampler in our BiMD. Since the backward flow cannot tune the starting step (image collapse), we double the manipulation scale of forward flow to simulate ablating half of the backward flow and then we tune the intervention starting  steps step of forward flow. 
Fig. \ref{qulitysteps} shows  examples with different MSS steps.  We can observe that  
personalized models do not have  ability to generate large-scale scene (smaller step), often tending to discard the scene and generate only face area.  The distribution learned by these personalized models is narrowed down to facial area. 
%
%
Our method integrates  these personalized models' generation process  with additional scene and facial information (larger step), helping  them access more information and expanding their generative capability.  
%

\begin{figure} [!ht] 
    \centering
    \includegraphics[width=0.23\textwidth]{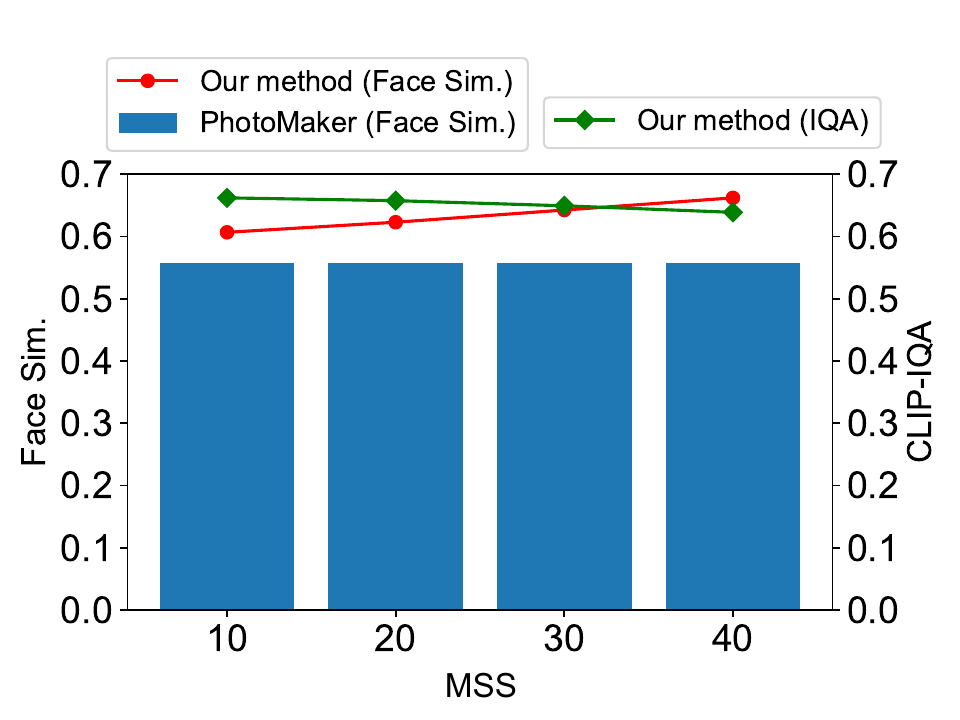} 
    \includegraphics[width=0.23\textwidth]{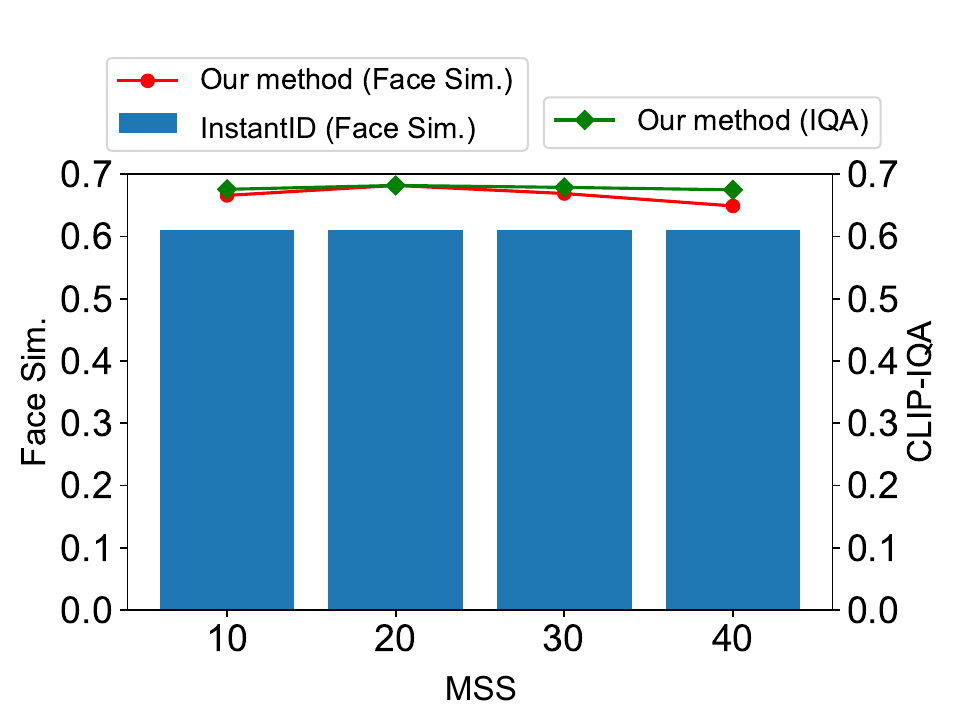} 
\caption{Face Sim. and CLIP-IQA across MSS for PhotoMaker and InstantID.} \label{bar}
\end{figure}

For better illustration, the quantitative results on face similarity and CLIP-IQA at different MSS are shown in  Fig. \ref{bar}. 
Results show that our method outperforms PhotoMaker and InstantID in all MSS ranges in terms of face similarity (Face Sim.). The increasing face similarity is   provided by representation from face swapping, 
which shows that  face swapping can supply additional identity information that neural network-based extractors cannot achieve. The image quality varies among different MSSs, but the variance are not significant.


\section{Application}

\begin{figure} [!ht]
    \centering 
    \includegraphics[width=0.42\textwidth]{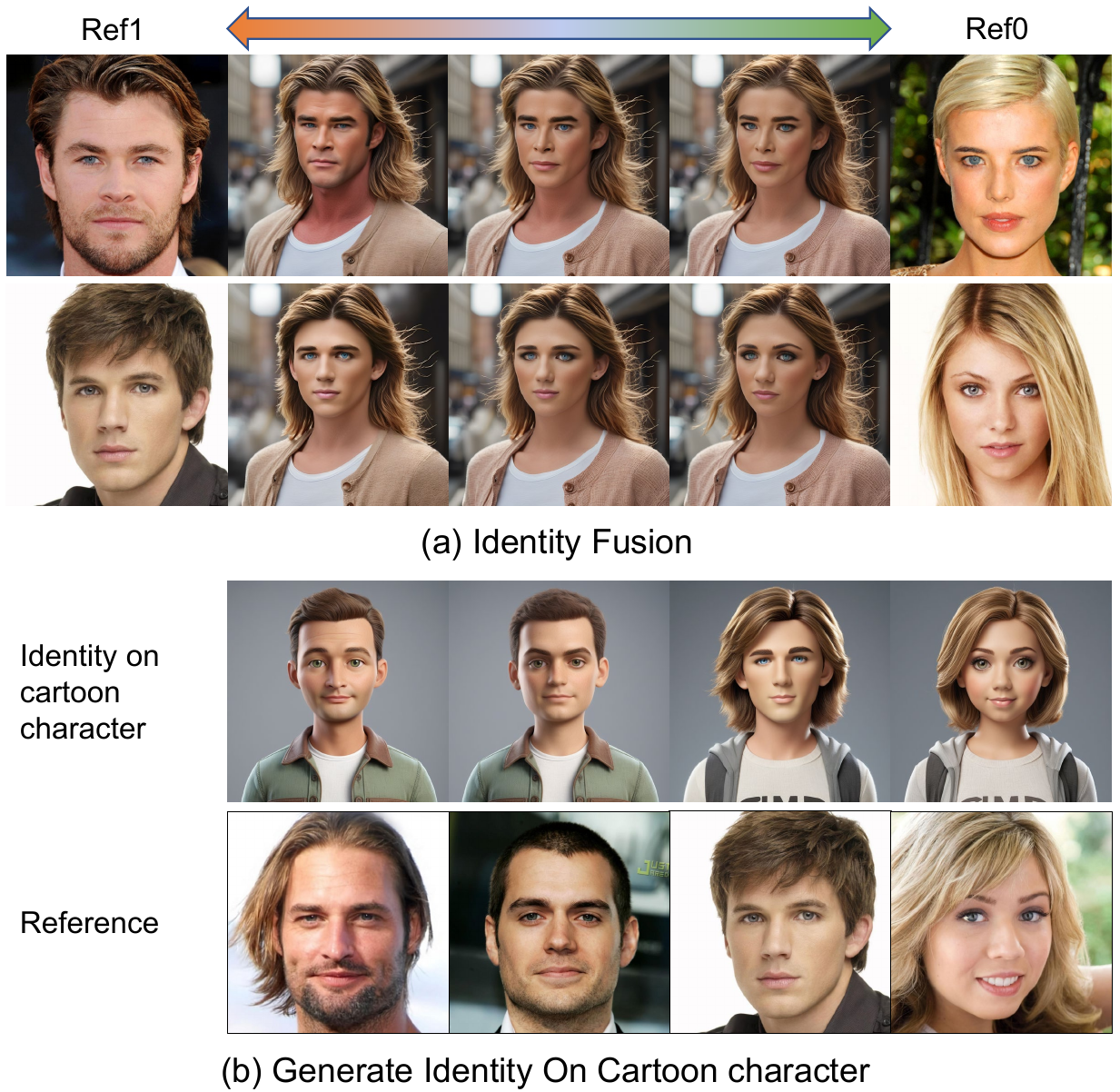} 
    \caption{
    Two applications of CustomEnhancer: (a) Identity fusion: Interpolate between two identities to do the identity synthesis. (b) Identity on specific cartoon character: Synthesis ID into specific character. 
    } \label{id_fusion}
\end{figure}

\paragraph{Identity fusion} 
Our pipeline implements  forward ID and backward ID two passes, which enables the  interpolation of two ID 
images.   
This allows for performing ID fusion  across the two IDs. Furthermore, it is able to provide explicit visualization of ID transformation trajectories through these two IDs, 
which PhotoMaker and InstantID cannot. 
We employ MSS to control the strength of forward ID, and we keep the manipulation scale of forward and backward flows' consistency. 
The results are provided in  Fig. \ref{id_fusion} (a). 
We can clearly observe how the identity is progressively fused through such an interpolation process.

\begin{figure}[!ht]
  \centering  \includegraphics[width=0.5\textwidth]{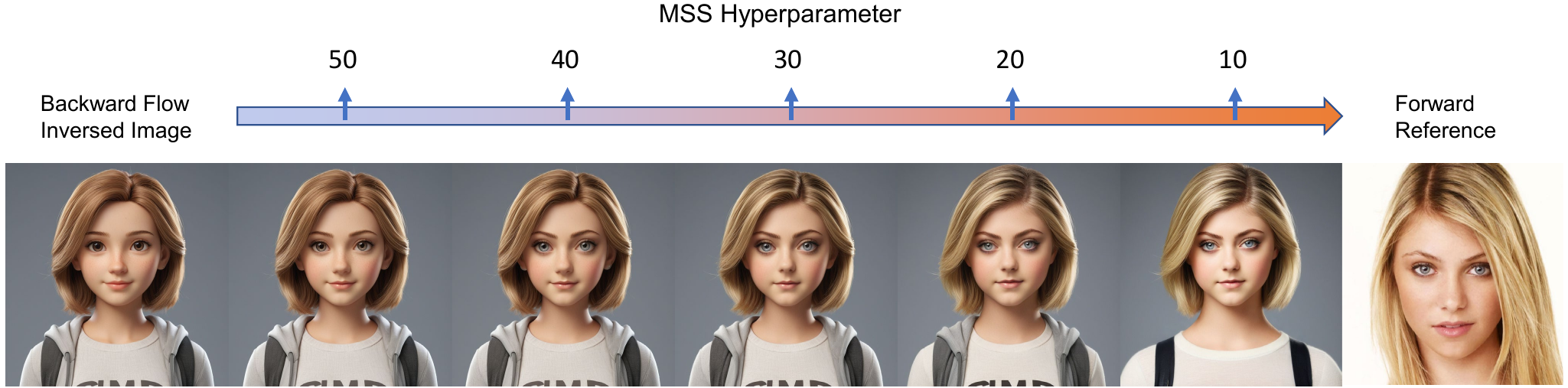}
  \caption{
Interpolation of identity transformation from provided specific cartoon character  to the reference  image. 
The first column shows the image inverted to the personalized  model through the backward flow, which is an ID-agnostic cartoon character. The last column represents the reference image fed forward into the personalized model. 
  }
  \label{appendix_cartoon_interpolation}
\end{figure}

\paragraph{Identity on specific cartoon character}

Our pipeline enables generating  identities on arbitrary specific cartoon characters, such as 2D  and sketches, through our unifying  generation process of BiMD.  The results are illustrated in Fig. \ref{id_fusion} (b).  Results show identity modification 
focused on semantic area of the head of specific cartoon character, allowing adjustments to the hairstyle, expression, etc. Furthermore, it can provide explicit visualization of transformations for these adjustments, 
as shown in Fig. \ref{appendix_cartoon_interpolation}, enabling  novel applications such as the discovery of unseen facial expressions, among others, etc.


\section{Conclusion}

In this paper, we  propose the CustomEnhancer for personalized photo generation models enhancing the scene diversity, training-free controls, and perceptual identity. We propose a zero-shot enhancement pipeline, BiMD approach,  ResInversion, and training-free controls, 
improving scene diversity by $6\%\sim8\%$, face similarity  by $7\%\sim9\%$, inversion speed $65\times\sim71\times$, offering controlled personalization without inefficient controller re-training.












\bibliographystyle{IEEEtran}
\bibliography{aaai2026.bib}

\vfill

\end{document}